\documentclass[10pt,twocolumn,letterpaper]{article}

\usepackage{iccv}
\usepackage{times}
\usepackage{epsfig}
\usepackage{graphicx}
\usepackage{amsmath}
\usepackage{amssymb}

\usepackage{booktabs}
\usepackage{caption}
\usepackage{subcaption}
\usepackage{multirow}
\usepackage{comment}
\usepackage{diagbox}

\usepackage{colortbl}  
\usepackage{xcolor}
\usepackage{array} 

\usepackage[breaklinks=true,bookmarks=false]{hyperref}

\iccvfinalcopy 

\def\iccvPaperID{****} 

\ificcvfinal\fi

\def \ours {ES-MVSNet}

\begin{document}

\title{\ours{}: Efficient Framework for End-to-end Self-supervised Multi-View Stereo}

\author{Qiang Zhou\\
{\tt\small mightyzau@gmail.com}
\and
Chaohui Yu\\
{\tt\small huakun.ych@alibaba-inc.com}
\and
Jingliang Li\\
{\tt\small lijingliang20@mails.ucas.ac.cn}
\and 
Yuang Liu \\
{\tt\small frankliu624@gmail.com}
\and 
Jing Wang \\
{\tt\small yunfei.wj@alibaba-inc.com}
\and 
Zhibin Wang \\
{\tt\small zhibin.waz@alibaba-inc.com}
}

\maketitle
\ificcvfinal\thispagestyle{empty}\fi

\begin{abstract}
Compared to the multi-stage self-supervised multi-view stereo (MVS) method, the end-to-end (E2E) approach has received more attention due to its concise and efficient training pipeline. Recent E2E self-supervised MVS approaches have integrated third-party models (such as optical flow models, semantic segmentation models, NeRF models, etc.) to provide additional consistency constraints, which grows GPU memory consumption and complicates the model's structure and training pipeline.
In this work, we propose an efficient framework for end-to-end self-supervised MVS, dubbed \ours{}.  To alleviate the high memory consumption of current E2E self-supervised MVS frameworks, we present a memory-efficient architecture that reduces memory usage by 43\% without compromising model performance. Furthermore, with the novel design of asymmetric view selection policy and region-aware depth consistency, we achieve state-of-the-art performance among E2E self-supervised MVS methods, without relying on third-party models for additional consistency signals.
Extensive experiments on DTU and Tanks\&Temples benchmarks demonstrate that the proposed \ours{} approach achieves state-of-the-art performance among E2E self-supervised MVS methods and competitive performance to many supervised and multi-stage self-supervised methods.
\end{abstract}


\section{Introduction}


\begin{figure}[!t]
    \centering
    \includegraphics[width=1.0\linewidth]{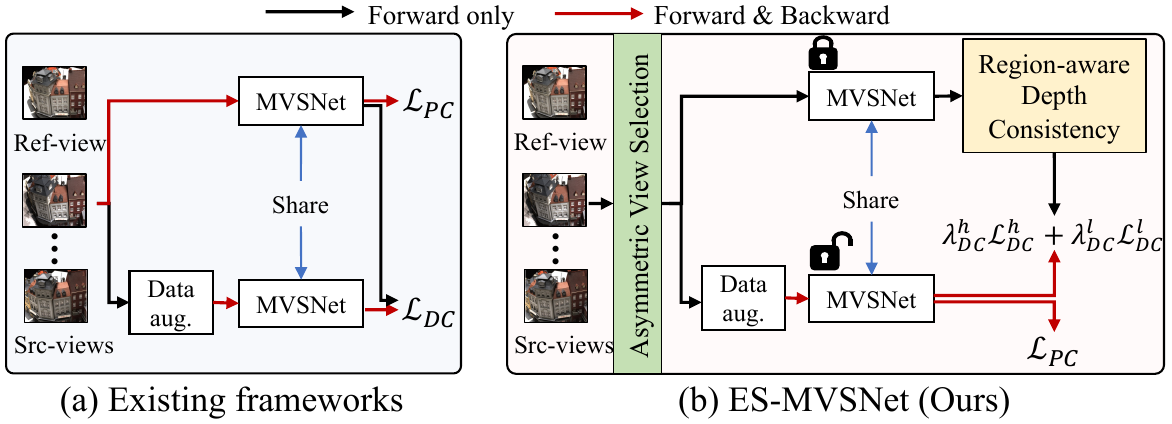}
    \vspace{-.25in}
    \caption{High-level comparison between \ours{} and existing E2E self-supervised MVS frameworks~\cite{jdacs_2021,rcmvsnet-2022,umvsnet_21}.
    $\mathcal{L}_{PC}$, $\mathcal{L}_{DC}$ denote photometric and depth consistency losses, respectively.}
    \label{fig:baseline_framework}
    \vspace{-.2in}
\end{figure}

Multi-View Stereo (MVS)~\cite{mvs_2016} is a long-standing fundamental task in 3D computer vision, aiming to recover 3D point clouds of real scenes from multi-view images and corresponding calibrated cameras. 
Like on other vision tasks, deep learning has driven the rapid development of MVS in recent years. Especially on public datasets like DTU~\cite{dtu_dataset_2016} and Tanks\&Temples~\cite{tanks_2017}, the end-to-end MVS depth estimation models~\cite{mvsnet_18,rmvsnet_2019,cascademvs_2020,aa-rmvsnet_2021,unimvsnet_2022} significantly improve the reconstruction performance compared with traditional geometry-based methods~\cite{colmap_1_2016,colmap_2_2016}.
MVSNet~\cite{mvsnet_18} is one of the representative MVS models based on fully supervised learning, which proposes to encode RGB information from different camera views into a cost volume, then predicts a depth map for point cloud reconstruction. The following supervised approaches~\cite{cascademvs_2020,rmvsnet_2019,aa-rmvsnet_2021,gbinet_22,unimvsnet_2022} improve the neural network architecture, reduce memory usage, and acquire state-of-the-art depth estimation performance on multiple benchmarks.
However, acquiring ground-truth depth data for supervision is difficult and expensive, limiting the practical use of these MVS models in general real-world scenarios. To reduce the reliance on ground-truth depth data, recently proposed self-supervised MVS methods~\cite{jdacs_2021,rcmvsnet-2022,kd-mvs-22} have received increasing attention.

Among the self-supervised MVS methods, the end-to-end (E2E) methods draw more attention due to their concise framework and efficient training.
The commonly used framework is shown in Figure~\ref{fig:baseline_framework}(a), in which the photometric consistency loss is applied in the weak-augmentation branch, and the depth consistency loss is applied in the strong-augmentation branch.
The photometric consistency assumes that pixels belonging to the same 3D point have the same color properties in different views. 
The depth consistency leverages the predicted depth maps of the weak-augmentation branch to supervise the predictions of the strong-augmentation branch.
To further improve performance, recent works introduce other consistency signals from third-party models, such as optical flow, segmentation, and NeRF models~\cite{umvsnet_21,jdacs_2021,rcmvsnet-2022}.
The E2E self-supervised MVS framework consumes vast GPU memory due to the 3D-CNN for cost volume regularization.
When additional third-party models are further introduced, the requirements for memory usage become 
more stringent, which significantly limits the application of these methods in high-resolution scenarios.

\begin{figure}[!t]
    \centering
    \includegraphics[width=0.99\linewidth]{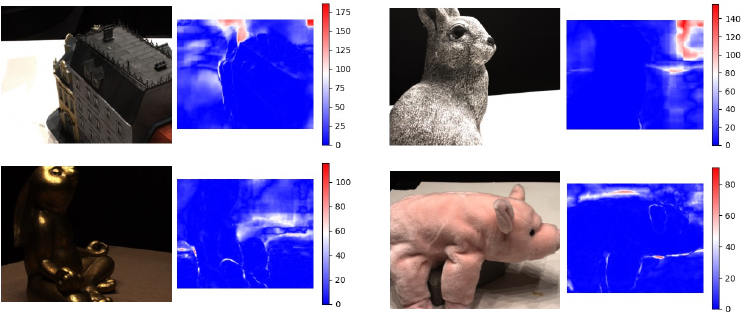}
    \vspace{-.1in}
    \caption{Imbalance in depth consistency loss. Some regions, such as the background and object boundaries, have a large loss value and may overwhelm others.}
    \label{fig:unbalance_depth_loss}
    \vspace{-.1in}
\end{figure}

In this work, we propose an efficient E2E self-supervised MVS framework that does not introduce third-party models and achieves state-of-the-art performance on DTU~\cite{dtu_dataset_2016} and Tanks\&Temples~\cite{tanks_2017} datasets.
\textbf{Our first contribution} is to propose a memory-efficient structure, as shown in Figure~\ref{fig:baseline_framework}(b). 
In our framework, photometric consistency and depth consistency losses are applied to the strong-augmentation branch, leaving the parameters of the weak-augmentation branch frozen.
The weak-augmentation branch only needs to infer the pseudo-depth map without the backward operation required, which reduces the GPU memory usage by 43\%.
Furthermore, the model performance improves when the photometric consistency is applied to more diverse augmented samples.
%
%
%
\textbf{Our second contribution} is to improve the efficacy of depth consistency. Specifically, we propose an asymmetric view selection policy. 
For the weak-augmentation branch, as in existing work, the top-$K$ source views are selected according to view-scores~\cite{mvsnet_18} to improve the accuracy of predicted pseudo-depth maps.
For the strong-augmentation branch, we randomly select source views according to the view-scores to increase the view diversity between these two branches.
%
Furthermore, we observe an apparent imbalance in the depth consistency loss among different regions, as shown in Figure~\ref{fig:unbalance_depth_loss}, implying that using a single global loss weight is 
inadequate. 
We propose region-aware depth consistency to alleviate this issue, which 
partitions pseudo-depth maps into high-quality and low-quality regions via online cross-view checking and assigns them different loss weights.
Figure~\ref{fig:losses_visualize} shows that the loss values of the two partitions differ by order of magnitude, verifying the necessity of our proposed region-aware loss weights and the effectiveness of the partitioning scheme based on online cross-view checking.
The detailed structure of our framework is depicted in Figure \ref{fig:framework}.

Our method, dubbed \ours{}, achieves state-of-the-art point cloud reconstruction results in the competitive DTU benchmark~\cite{dtu_dataset_2016}, and also demonstrates robust performance on out-of-distribution samples in the Tanks\&Temples dataset~\cite{tanks_2017}.
In summary, our contributions are as follows:
\begin{itemize}
\itemsep -0.1cm
    \item We propose an efficient framework for end-to-end self-supervised MVS, dubbed \ours{}.
    \item We propose a memory-efficient architecture that reduces memory usage by 43\% without compromising model performance, which benefits the use in high-resolution scenarios. 
    \item We improve the efficacy of depth consistency through two novel designs: asymmetric view selection policy and region-aware depth consistency.
\end{itemize}

\section{Related Work}

\subsection{Supervised MVS}

Since the era of deep learning, many supervised learning-based models have been proposed to reconstruct 3D scenes~\cite{surfacenet_17,surfacenetplus_2021,mvsnet_18,cascademvs_2020,rmvsnet_2019,aa-rmvsnet_2021,unimvsnet_2022,gbinet_22}. Among these methods, 
depth map reconstruction demonstrates the most versatility, as it decouples the intricate MVS problem into a per-view depth map estimation problem.
One of the representative works is MVSNet~\cite{mvsnet_18}, which encodes camera parameters and backbone features into a cost volume via homography warping, and regularizes the volume via a 3D CNN before predicting the final depth map.
The 3D CNN in MVSNet consumes high GPU memory, which limits its usage in high-resolution scenes. To alleviate the memory usage problem, some works~\cite{aa-rmvsnet_2021,rmvsnet_2019,YanWYDZCWT20} propose replacing 3D CNNs with convolutional recurrent GRU~\cite{GRU2014} or LSTM~\cite{LSTM97} units.
The subsequent multi-stage architectures~\cite{cascademvs_2020,cvp-mvsnet-20,unimvsnet_2022} dramatically improve the model performance by learning depth predictions in a coarse-to-fine manner. The multi-stage architecture also balances model performance and memory cost better, where more depth hypotheses are employed in low-resolution stages and fewer in high-resolution stages.

However, the reliance on ground truth depth data limits the application of supervised MVS methods to a broader range of realistic scenarios. Therefore, it is essential to explore alternative self-supervised methods.

\subsection{Multi-Stage Self-supervised MVS}

Multi-stage self-supervised MVS usually first trains a teacher model based on photometric consistency. The teacher model is then used to generate pseudo-depth maps to supervise the training of the student model. 
Several cycles may be iterated to improve the performance of the student model, i.e., the previous student model acts as a new teacher model to guide the learning of the new student model.
For example, U-MVSNet~\cite{umvsnet_21} consists of two stages: self-supervised pre-training and pseudo-label-based post-training. In the first pre-training stage, a flow-depth consistency loss is introduced 
in addition to the photometric consistency loss. In the second post-training stage, the student model is supervised with pseudo-labels. 
CVP-MVSNet~\cite{cvp-mvsnet-20} proposes to learn an initial pseudo-depth map through unsupervised pre-training, then use a well-designed pipeline to refine the initial pseudo-depth map, and finally use the refined pseudo-depth map to supervise the training of the student model.
The KD-MVS~\cite{kd-mvs-22} first trains a teacher model in a self-supervised manner using photometric and feature consistency, then distills the knowledge from the teacher model to the student model via proposed probabilistic knowledge transfer.

Although multi-stage self-supervised MVS methods achieve superior performance, their complex framework and training process make it inconvenient to use in practical applications.

\subsection{End-to-end Self-supervised MVS}

Unlike the multi-stage approaches, the end-to-end approaches only train one model, making the training process more efficient and concise.
Unsup\_MVS~\cite{Khot_2019} proposes the first end-to-end self-supervised MVS framework, guiding the depth prediction by minimizing the discrepancy between the reference image and the inversely warped images from source views. The training objectives of photometric consistency, SSIM~\cite{SSIM_2004} loss, and depth smoothing used in this work are widely used in subsequent self-supervised work.
JDACS~\cite{jdacs_2021} incorporates data augmentation into self-supervised MVS and applies depth consistency to constrain the prediction of weak and strong augmentation branches.
It further proposes cross-view semantic consistency implemented via an unsupervised co-segmentation module.
%
To alleviate the correspondence ambiguity among views due to occlusion, etc., RC-MVSNet~\cite{rcmvsnet-2022} adds an additional NeRF~\cite{nerf_2020} branch, which shares the backbone with the original MVS branch, and imposes rendering consistency on the depth maps predicted by these two branches.

In this work, instead of introducing additional consistency signals with third-party models such as segmentation and NeRF, we focus on improving the efficacy of depth consistency and propose a memory-efficient framework.

\begin{figure*}
    \centering
    \includegraphics[width=0.99\linewidth]{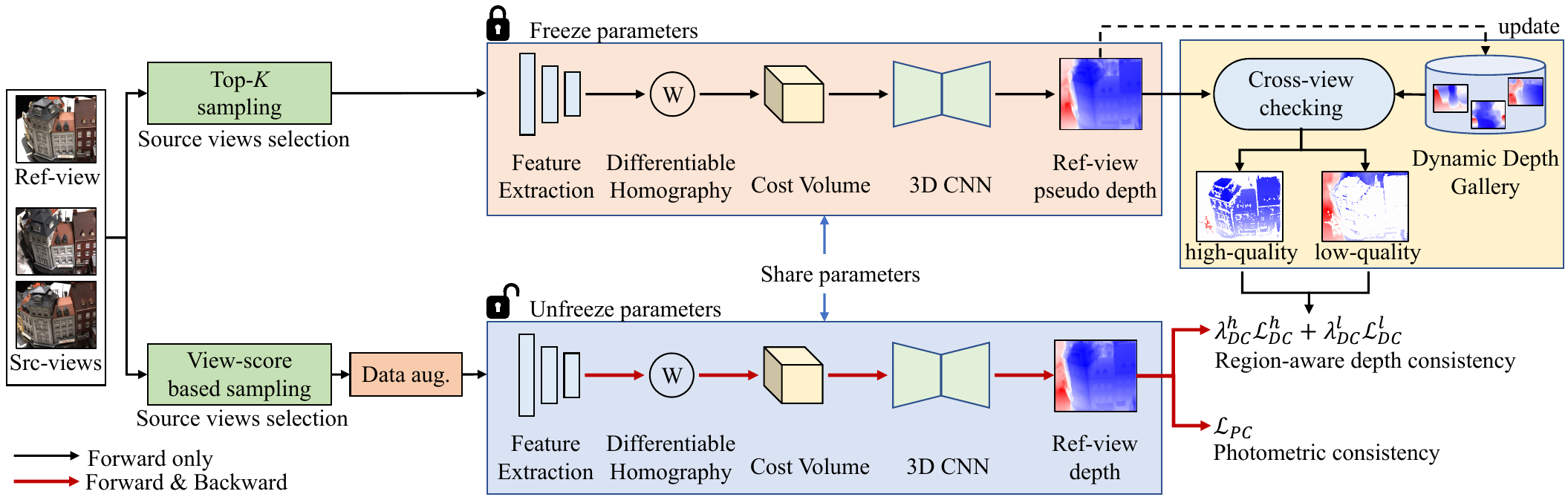}
    \vspace{-.05in}
    \caption{Overview of our proposed framework (\ours{}) for end-to-end self-supervised MVS.
    Compared with previous methods, our proposed approach possesses three novel designs: memory-efficient design, asymmetric view selection strategy, and region-aware depth consistency.}
    \label{fig:framework}
    \vspace{-.15in}
\end{figure*}

\section{Method}

In this section, we describe \ours{}, our proposed approach for end-to-end self-supervised multi-view stereo.
Given $N$ images as input (by default, $1st$ is the reference view), with their corresponding cameras' intrinsic and extrinsic parameters, our method predicts a depth map in the reference camera view.
The overall pipeline is illustrated in Figure~\ref{fig:framework}.
In the following, we first describe the existing baseline framework in Section~\ref{sec:basline}. Then we elaborate innovative designs of our approach, including memory-efficient design (Section~\ref{sec:memory-efficient}), asymmetric view selection policy (Section~\ref{sec:view_policy}), and region-aware depth consistency (Section~\ref{sec:balance_depth_consistency}).

\subsection{Preliminary}
\label{sec:basline}
This section introduces the model framework and training loss used in existing end-to-end self-supervised MVS work~\cite{jdacs_2021,rcmvsnet-2022,umvsnet_21}. 
Third-party models such as optical flow, segmentation, and NeRF models are excluded for providing a clean baseline.

\paragraph{Model Architecture.}
The framework in Figure~\ref{fig:baseline_framework}(a) is adopted by recent E2E self-supervised MVS methods, which contains two branches of weak-augmentation and strong-augmentation respectively.
The two branch models have the same structure and share model parameters.
The weak-augmentation branch takes the original image as input and finally applies a photometric consistency loss on the predicted depth map. The strong-augmentation branch applies color perturbation to the input image to simulate color inconsistency between views, and finally applies a depth consistency loss to the predicted depth map.
For each branch, the network firstly extracts features using a CNN from $N$ input images. Then a variance-based cost volume~\cite{mvsnet_18,AdaptiveViewAggregation_2020} is constructed via differentiable homography warping and a 3D U-Net is used to regularize the 3D cost volume. Finally, the depth map is inferred for every reference image.

\paragraph{Overall Loss.}

The photometric consistency loss is applied to the weak-augmentation branch to minimize the difference between the warped image and the original image at the same view.
For a particular image pair $(I_1, I_j)$ with associated intrinsic and extrinsic parameters $(K, T)$, we can calculate the corresponding projected location $p'_j$ in the source view from its coordinates $p_1$ in the reference view:
\begin{equation}
    p'_{j,k} = K_j T_{1j} (D_1(p_{1, k}) K_1^{-1} p_{1, k}),
\end{equation}
where $k (1 \leq k \leq HW)$ is the index of the pixels, $H$ and $W$ denote the images' height and width, and $D_1$ represents the predicted depth map in the reference view.
Through differentiable bilinear sampling, we can obtain the warped image $I_1^j$ under the reference view, taking the source view image $I_j$ and the predicted depth map $D_1$ as input.
Along with the warping, a binary validity mask $M_j$ is generated simultaneously, indicating whether the projected position $p'_j$ 
lies within the valid image region.
During training, all $N-1$ source views are warped to the reference view to compute the photometric loss:

\begin{footnotesize}
\begin{equation}
    \mathcal{L}_{PC} = \sum_{j=2}^{N}\frac{\left \| (I_1^j - I_1) \odot M_j \right \|_2 + \left \| (\nabla I_1^j - \nabla I_1) \odot M_j \right \|_2  }{\left \| M_j \right \|_1 },
\end{equation}
\end{footnotesize}
where $\odot$ denotes Hadamard product, $\nabla$ represents the gradient of pixels.

The depth consistency loss is applied to the predicted depth maps $D_{1}^s$ of the strong-augmentation branch, taking the output depth maps $D_1^w$ of the weak-augmentation branch as pseudo ground truths.

\begin{equation}
    \mathcal{L}_{DC} = \frac{1}{HW}\sum^{HW}_{k=1} \left \| D^s_{1,k} - D^w_{1,k} \right \|_2.
\label{eq:depth_consistency}
\end{equation}

The final training objective can be constructed as follows:
\begin{equation}
\begin{aligned}
    \mathcal{L} = &\lambda_{PC} \mathcal{L}_{PC} + \lambda_{DC} \mathcal{L}_{DC} \\
            &+ \lambda_{SSIM} \mathcal{L}_{SSIM} + \lambda_{Smooth} \mathcal{L}_{Smooth},
\end{aligned}
\label{eq:losses_baseline}
\end{equation}
where two commonly used regularization terms for depth estimation are also applied, including structural similarity $\mathcal{L}_{SSIM}$~\cite{SSIM_2004} and depth smoothness $\mathcal{L}_{Smooth}$~\cite{RadfordKHRGASAM_2021, Khot_2019}.
The weights are empirically set as: $\lambda_{PC} = 0.8$, $\lambda_{DC} = 0.1$, $\lambda_{SSIM} = 0.2$, $\lambda_{Smooth} = 0.0067$.

\subsection{Memory-efficient Design}
\label{sec:memory-efficient}

One major limitation of deep learning based MVS methods is scalability: 
%
the memory-intensive 3D-CNN renders the learned MVS difficult to apply to high-resolution scenarios.
This limitation is especially pronounced in the current self-supervised MVS framework~\cite{jdacs_2021,rcmvsnet-2022,umvsnet_21}, which almost doubles memory usage. As shown in Figure~\ref{fig:baseline_framework}(a), the network needs to perform forward inference and reverse gradient calculation for both the weak and strong augmentation branches. 
In this work, we redesign the E2E self-supervised MVS framework, which effectively reduces memory consumption and improves the model performance.

Formally, we freeze the weak-augmentation branch and apply photometric consistency and depth consistency to the output of the strong-augmentation branch, as shown in Figure~\ref{fig:framework}.
The weak-augmentation branch only conducts forward operations to prepare pseudo-depth maps, reducing GPU memory usage by 43\%, from 14100 MiB to 8000 MiB
as reported in Table~\ref{tbl:memory_efficient}.
The results in Table~\ref{tbl:memory_efficient} also reveal that applying photometric consistency to the strong-augmentation branch benefits model performance, improving the overall metric from 0.3483 to 0.3371 on the DTU dataset.
The considerable performance gain mainly comes from supplying more diverse augmentation samples for photometric consistency. 
Another experiment also confirms the efficacy of providing data augmentation samples for photometric consistency, as shown in Table~\ref{tbl:aug_photometric}, where the interference of depth consistency is excluded.

\subsection{Asymmetric View Selection Policy}
\label{sec:view_policy}

\begin{figure}[!t]
    \centering
    \includegraphics[width=0.99\linewidth]{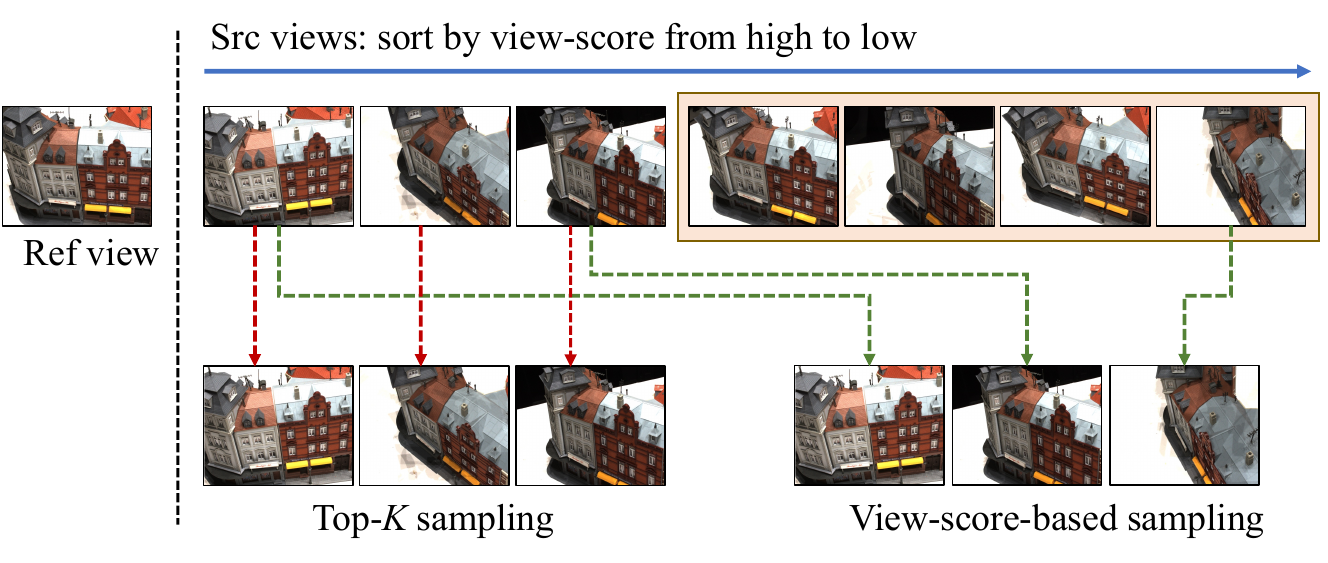}
    \vspace{-.1in}
    \caption{
    Depiction of view-score-based sampling.
    Compared to the top-$K$ sampling, our view-score-based sampling can select any source view sample, with a higher probability of selecting views with high view-scores.
    }
    \label{fig:view_policy}
    \vspace{-.15in}
\end{figure}

Several source views $I_{2 \rightarrow N}$ are selected to provide geometric information when predicting the depth map of the reference view $I_1$.
Following MVSNet, recent self-supervised MVS methods select source views with top-${(N-1)}$ view-scores~\cite{mvsnet_18} for the weak and strong augmentation branches.
We name this the top-$K$ view selection policy.
The top-$K$ policy selects those source views with the optimal viewpoint difference from the reference view, leading to more accurate predicted depth maps.
However, sharing the same source view selection policy on weak and strong branches limits the sample 
diversity
between the two branches.

In this work, we explore an asymmetric view selection policy to leverage weak-strong augmentation branches more efficiently and
boost performance.
Specifically, for the weak-augmentation branch, we maintain the top-$K$ view selection policy to provide accurate pseudo-depth maps. For the strong-augmentation branch, we propose a view-score-based selection policy.
%
%
The view-score-based policy is designed based on two considerations. First, compared with the top-$K$ policy, our approach can select any source view. In this way, for a given reference view, the scope of the depth consistency constraint extends from color augmentation samples to viewpoint diversity samples. Second, compared with the random sampling policy, we assign higher selection probabilities to source views with higher view-scores, making training more stable and achieving higher performance.
Experiments in Table~\ref{tbl:view_policy} verify that the random sampling policy improves performance by introducing more diverse source view samples, while our view-score-based policy acquires the best performance.
%
Figure~\ref{fig:view_policy} depicts our view-score-based sampling.
To be specific, for the strong-augmentation branch, the $N-1$ source views are selected from all source views according to their view-scores (i.e., views with higher view-scores have a greater probability of being selected), where the view-score takes the definition as MVSNet~\cite{ZhangLFZQ15,mvsnet_18}.


\subsection{Region-aware Depth Consistency}
\label{sec:balance_depth_consistency}

\begin{figure}[!t]
    \centering
    \includegraphics[width=0.95\linewidth]{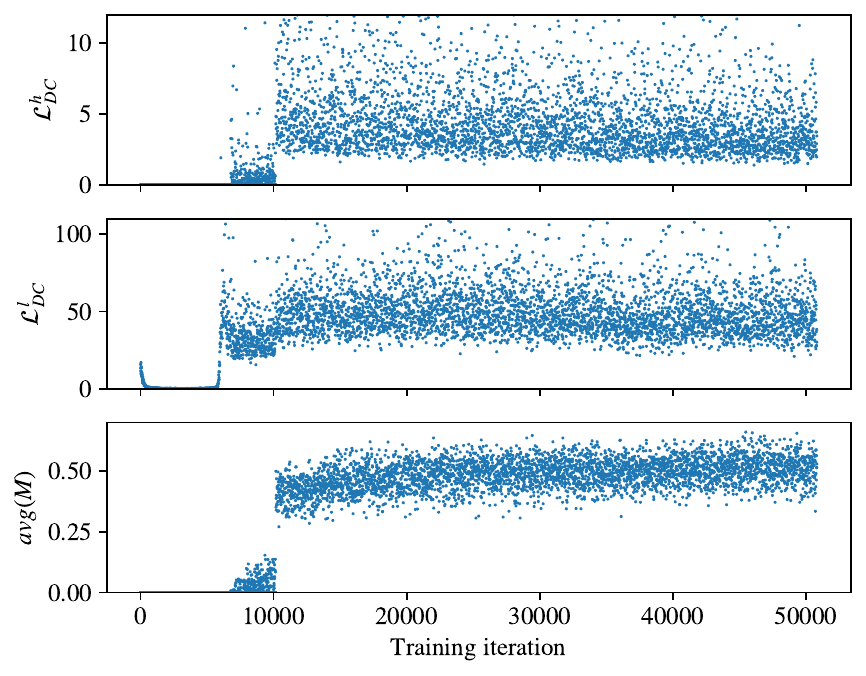}
    \vspace{-.1in}
    \caption{
    Partition visualization for depth consistency loss based on online cross-view checking.
    $\mathcal{L}_{DC}^h$ denotes the depth consistency loss for high-quality regions. $\mathcal{L}_{DC}^l$ indicates depth consistency in low-quality regions. $avg(M)$ represents the ratio of high-quality pseudo-depths in the entire pseudo-depth map.
}
    \label{fig:losses_visualize}
    \vspace{-.15in}
\end{figure}

Depth consistency plays an important role in end-to-end self-supervised MVS methods, where the output of the weak-augmentation branch is employed as the pseudo-depth map to supervise the learning of the strong-augmentation branch.
However, as shown in Figure~\ref{fig:unbalance_depth_loss}, the depth consistency loss suffers from imbalance among pixels, where texture-less backgrounds or object boundaries (with sudden changes in depth) usually lead to a significant increase in depth consistency loss.
These regions usually correspond to erroneous pseudo-depths where predictions are inaccurate. 
U-MVSNet~\cite{umvsnet_21} selects to filter out those unreliable pseudo-depths through the uncertainty maps obtained by Monte-Carlo Dropout~\cite{monte_carlo_2017}.
In contrast, as shown in Table~\ref{tbl:abl_region_aware}, we experimentally find that even those erroneous pseudo-depths play a significant positive role in depth consistency. A more reasonable solution is to keep accurate and erroneous pseudo-depths, while each employs different loss weights to alleviate the imbalance in depth consistency.
In this work, we propose alleviating the imbalance in depth consistency via region-aware loss weights.

%
%
Formally, as shown in Figure~\ref{fig:framework}, we use a dynamic depth gallery to cache and update pseudo-depth maps for all reference views during the training phase. These cached pseudo-depth maps are then utilized to separate the pseudo-depth maps into high-quality and low-quality regions with online cross-view checking.
%
%
Taking an arbitrary pixel $\mathbf{p}_{1,k}$ in the reference view as an example,
we first cast the 2D point to the corresponding 3D point $\mathbf{P}_{1,k}$ with the pseudo-depth value $D_{1}^w(\mathbf{p}_{1,k})$ (from the prediction of the weak-augmentation branch).
Then, we project the 3D point to the $i$-th source view with the camera intrinsics and extrinsics, getting the 2D point $\mathbf{p}_{i,k}$.
Finally, with the pseudo-depth of $D_i^w(\mathbf{p}_{i,k})$ (from cached depth maps), we re-project the 2D point $\mathbf{p}_{i,k}$ back to the reference view, and getting the projected 2D point $\mathbf{\hat{p}}_{1,k}$ and depth value $\hat{D}_1(\mathbf{\hat{p}}_{1,k})$ in the reference view. 
%
By defining the re-projection error in pixels and depth as $e_{pixel}=\left \| \mathbf{p}_{1,k} - \mathbf{\hat{p}}_{1,k} \right \|_2$ and $e_{depth} = \left \| \hat{D}_1(\mathbf{\hat{p}}_{1,k}) - D_1^w(\mathbf{p}_{1,k}) \right \|_1 / D_1^w(\mathbf{p}_{1,k}) $,
the high-quality mask $M_1$ of the reference view is computed as
\begin{equation}
\begin{aligned}
    M_{1, k} &= (C_{1,k}^w > \tau_1) \\
        &\left [ \left (\sum_{i=2}^{S} \left ( e_{pixel} < \tau_2) \cdot (e_{depth} < \tau_3   \right ) \right ) >= \tau_4 \right ],
\end{aligned}
\label{eq:quality_mask}
\end{equation}
where $k$ is the index of the pixel, $C_1^w$ is the predicted confidence map, and $S$ is the total number of source views (10 by default).
Hyperparameters $\tau_1$, $\tau_2$, $\tau_3$, and $\tau_4$ are default set to 0.5, 0.5, 0.01, and 4.

With the high-quality mask computed in Equation~\ref{eq:quality_mask}, the pseudo-depth map $D_1^w$ is partitioned into high-quality region $D_1^{w,h}=D_1^w \odot M_1$ and low-quality region $D_1^{w,l} = D_1^w \odot (1 - M_1)$, and we reformulate the training loss of Equation~\ref{eq:losses_baseline} as
\begin{equation}
\begin{aligned}
    \mathcal{L} = &\lambda_{PC} \mathcal{L}_{PC} + \lambda_{DC}^h \mathcal{L}_{DC}^h + \lambda_{DC}^l \mathcal{L}_{DC}^l \\ 
             &+ \lambda_{SSIM} \mathcal{L}_{SSIM} + \lambda_{Smooth} \mathcal{L}_{Smooth},
\end{aligned}
\label{eq:losses_new}
\end{equation}
where $\lambda_{DC}^h$ and $\lambda_{DC}^l$ denote the depth consistency loss weights for high-quality and low-quality regions, respectively.
%
As shown in Figure~\ref{fig:losses_visualize}, our proposed online cross-view checking can effectively partition depth consistency, where the consistency loss of partitioned low-quality regions $\mathcal{L}_{DC}^l$ is an order of magnitude higher than $\mathcal{L}_{DC}^h$ of high-quality regions.
Reasonably setting the loss weights $\lambda_{DC}^h$ and $\lambda_{DC}^l$ of these two regions will maximize the impact of depth consistency, which is the motivation of our proposed region-aware depth consistency.

\begin{figure*}[!t]
    \centering
    \includegraphics[width=0.75\linewidth]{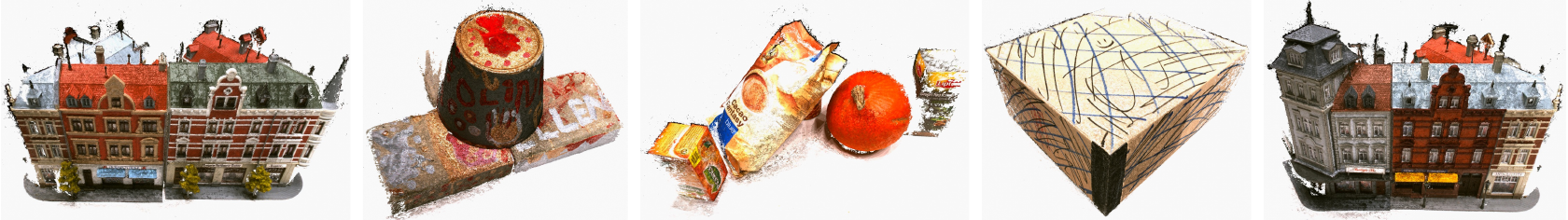}
    \caption{Qualitative results on some scenes of DTU dataset~\cite{dtu_dataset_2016}.}
    \label{fig:dtu_visualization}
\end{figure*}

\begin{figure*}[!t]
    \centering
    \includegraphics[width=0.75\linewidth]{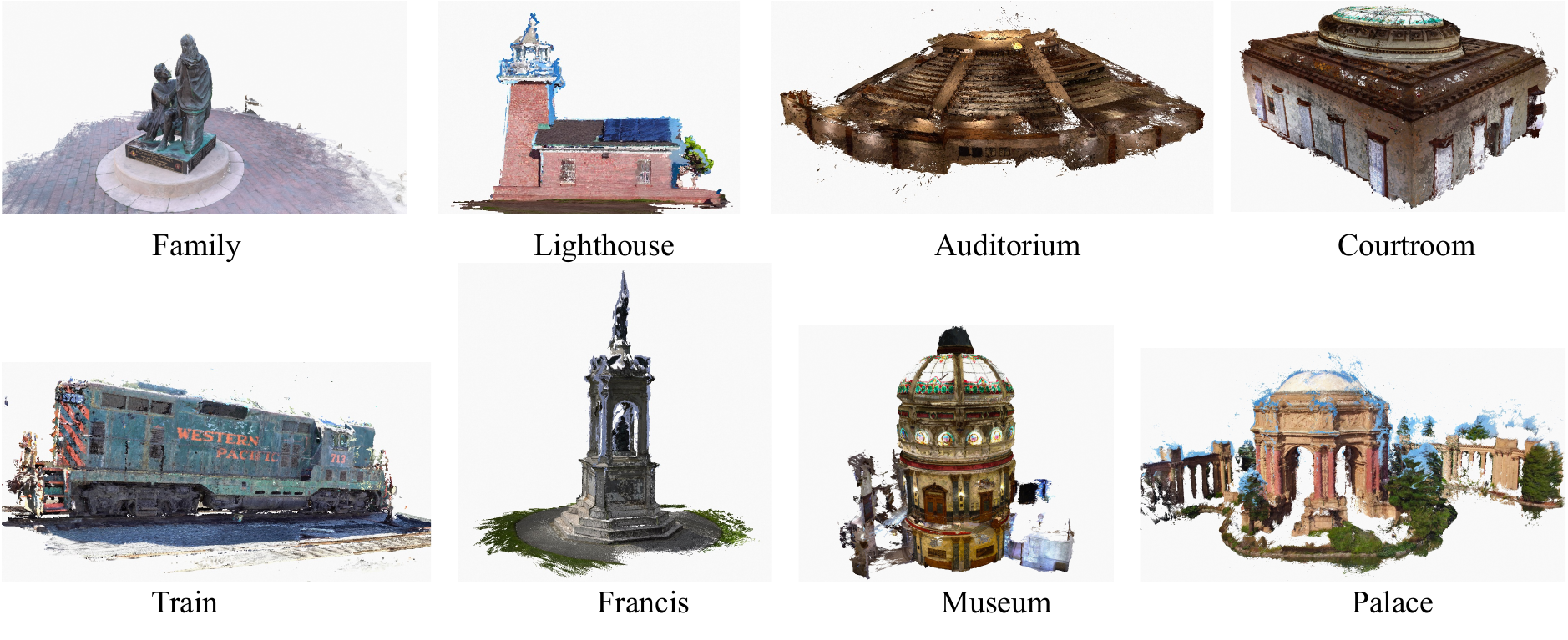}
    \vspace{-.1in}
    \caption{Qualitative results on scenes of Tanks\&Temples dataset~\cite{tanks_2017}. 
    The model is trained on the DTU training set only.}
    \label{fig:tanks_visualize}
    \vspace{-.1in}
\end{figure*}

\section{Experiments}

\subsection{Datasets}

\textbf{DTU}~\cite{dtu_dataset_2016} is an indoor dataset with multi-view images and corresponding camera parameters. There are 124 scenes scanned from 49 or 64 views under seven lighting conditions. We follow the setup of MVSNet~\cite{mvsnet_18} to divide the training, validation, and evaluation sets. In the DTU benchmark, the model is trained on the training set and tested on the evaluation set. 
We employ official error metrics from DTU to evaluate \textit{Accuracy}, \textit{Completeness}, and \textit{Overall}.

\textbf{Tanks\&Temples}~\cite{tanks_2017} is a large-scale dataset containing various outdoor scenes. It includes an intermediate subset and an advanced subset. This benchmark is evaluated online by submitting the generated point cloud to the official website. In this benchmark, the F-score is calculated for each scene, and we separately report the average F-scores for the intermediate and advanced subsets.

\subsection{Implementation Details}

\textbf{Training Details.} The proposed \ours{} is trained on the DTU~\cite{dtu_dataset_2016} dataset.
Following~\cite{MVS2_2019,M3VSNET_2021,Khot_2019,jdacs_2021,rcmvsnet-2022}, we use the high-resolution DTU data provided by the open
source code of MVSNet~\cite{mvsnet_18}. We first resize the input images to $600 \times 800$, following previous methods. Then we crop resized images into $512 \times 640$ patches.
The number of images $N$ is set to 5 and 4 for the weak and strong augmentation branches, respectively. Data augmentation strategies are the same as JDACS~\cite{jdacs_2021}, including gamma correction and color jitter.
We adopt the backbone of Cas-MVSNet~\cite{cascademvs_2020} to construct our multi-scale pipeline with 3 stages. For each stage, we use different feature maps and the 3D-CNN network parameters. The whole network is optimized by an Adam optimizer in Pytorch for 15 epochs with an initial learning rate of 0.0001, which is downscaled by a factor of 2 after 10, 12, and 14 epochs. We train with a batch size of 8 using eight NVIDIA V100 GPUs.

\textbf{Testing Details.} The model trained on DTU training set is used for testing on DTU testing set. The input image number $N$ is set to 5, each with a resolution of $1152 \times 1600$. 
The model trained on DTU training dataset is directly used for testing on Tanks\&Temples intermediate and advanced datasets without finetuning. The image sizes are set to $1024 \times 1920$ or $1024 \times 2048$ and the input image number $N$ is set to 7.
%
When fusing the per-view depth maps into the final point cloud, we employ the photometric and geometric consistencies, and the hyper-parameter settings are consistent with RC-MVSNet~\cite{rcmvsnet-2022}.

\subsection{Ablation Study}

\paragraph{Effect of memory-efficient design.}

\begin{table}[!t]
\centering
\resizebox{0.95\linewidth}{!}{
\begin{tabular}{@{}c|ccc|c@{}}
\toprule
M. design & Acc($\downarrow$)    & Comp($\downarrow$)   & Overall($\downarrow$) & GPU memory [MiB] \\ \midrule
   & 0.3762 & 0.3203 & 0.3483  & 14100    \\
$\checkmark$   & 0.3582 & 0.3160 & \textbf{0.3371}  & \hspace{.15in}8000 (\textcolor{red}{$\downarrow$})  \\ 
\bottomrule
\end{tabular}
}
\caption{Ablation study of memory-efficient design. ``M. design'' denotes memory-efficient design.
The loss weight $\lambda_{DC}$ is set to 0.1 by default.
}
\label{tbl:memory_efficient}
\end{table}

\begin{table}[!t]
\centering
\resizebox{0.75\linewidth}{!}{
\begin{tabular}{@{}c|c|ccc@{}}
\toprule
$\lambda_{DC}$             & Augmentation & Acc($\downarrow$)    & Comp($\downarrow$)   & Overall($\downarrow$) \\ \midrule
\multirow{2}{*}{0} &     None        &   0.3842 & 0.3510 & 0.3676  \\ 
                   &     Color       &   0.3594 & 0.3340 & \textbf{0.3467}   \\ 
\bottomrule
\end{tabular}
}
\caption{
The impact of data augmentation on photometric consistency.
The loss weight $\lambda_{DC}$ is set to 0 to exclude the affect of depth consistency.
}
\label{tbl:aug_photometric}
\end{table}

Our memory-efficient design freezes the parameters of the weak-augmentation branch and applies photometric consistency to the output of the strong-augmentation branch. As shown in Table~\ref{tbl:memory_efficient}, this small design reduces the memory usage of the model by 43\%, which is conducive to the application of the model in high-resolution scenarios. 
Table~\ref{tbl:memory_efficient} also reveals that the model performance is significantly improved, thanks to the photometric consistency applied on more diverse samples via data augmentation.
The additional experiments in Table~\ref{tbl:aug_photometric}, where the depth consistency is excluded by setting loss weight $\lambda_{DC}$ to 0, also confirm that applying photometric consistency to augmented samples helps improve the performance of self-supervised MVS.
In the following experiments, we employ the proposed memory-efficient framework by default.

\paragraph{Effect of asymmetric view selection policy.}

\begin{table}[!t]
\centering
\resizebox{0.99\linewidth}{!}{
\begin{tabular}{@{}c|c|ccc|c@{}}
\toprule
W. policy & S. policy  & Acc($\downarrow$)    & Comp($\downarrow$)   & Overall($\downarrow$) & Memory [MiB] \\ \midrule
\multirow{3}{*}{Top-$K$} & Top-$K$      &  0.3582 & 0.3160 & 0.3371 & 8000 \\
& Random             &    0.3525 & 0.3086 & 0.3306    &   8000  \\ 
& View-score-based              &  0.3520 & 0.3004 & \textbf{0.3262}   &  8000   \\ 
\midrule
View-score-based & View-score-based & 0.3545 & 0.2996 & 0.3271 & 8000 \\
\bottomrule
\end{tabular}
}
\caption{Ablation study of source view selection policy.
``W. policy'': the view selection policy for the weak-augmentation branch.
``S. policy'': the view selection policy for the strong-augmentation branch. 
The loss weight $\lambda_{DC}$ is set to 0.1 by default.
}
\label{tbl:view_policy}
\vspace{-.1in}
\end{table}

In this section, we investigate the impact of the asymmetric view selection policy. We first fix the view selection policy of the weak-augmentation branch as ``top-$K$'' to provide high-quality pseudo-depth maps. Then we vary the policy of the strong-augmentation branch. The experimental results in Table~\ref{tbl:view_policy} show that the model performance improves when switching the strong-augmentation branch's policy from ``top-$K$'' to ``random sampling''. Furthermore, the model performs best when adopting a sampling policy based on view-scores.
Compared with random sampling, the view-score-based sampling can effectively reduce the influence of those source views with significant viewpoint differences from the reference view.

\paragraph{Effect of region-aware depth consistency.}

\begin{table}[!t]
\centering
\resizebox{0.88\linewidth}{!}{
\begin{tabular}{@{}c|ll|ccc|c@{}}
\toprule
$\lambda_{DC}$ & $\lambda_{DC}^h$ & $\lambda_{DC}^l$ & Acc($\downarrow$)    & Comp($\downarrow$)   & Overall($\downarrow$) & GPU [MiB] \\ \midrule
0.1 &       &       & 0.3520 & 0.3004 & 0.3262 & 8000 \\ 
\midrule
& 0. & 0. & 0.3605 & 0.3234 & 0.3420 & 8000 \\
& 0.      & 0.1      & 0.3459  & 0.3086 & 0.3291 & 8000 \\ 
& 0.5 & 0. & 0.3550 & 0.2961 & 0.3255 & 8000 \\
& 0.5 & 0.1 & 0.3479 & 0.2983 & \textbf{0.3231} & 8000 \\
\bottomrule
\end{tabular}
}
\caption{Ablation study of region-aware depth consistency.
For the view selection policy, the strong-branch utilizes view-score-based sampling by default.
}
\label{tbl:abl_region_aware}
\vspace{-.1in}
\end{table}

\begin{table}[!t]
\centering
\resizebox{0.8\linewidth}{!}{
\begin{tabular}{@{}c|cccc@{}}
\toprule
\diagbox{$\lambda_{DC}^l$}{Overall($\downarrow$)}{$\lambda_{DC}^h$} & 0 & 0.1 & 0.5 & 1.0 \\ \midrule
0                    & 0.3420  & 0.3280 & 0.3255  &  0.3336   \\
0.1                  & 0.3291  & 0.3262  & \textbf{0.3231} &  0.3266  \\
0.2                  &  0.3329 &  0.3308 & 0.3275 & 0.3344   \\
0.4                  & 0.3481  & 0.3363  & 0.3357  & 0.3370 \\ \bottomrule
\end{tabular}
}
\caption{Detailed ablation of loss weights $\lambda_{DC}^h$ and $\lambda_{DC}^l$.}
\label{tbl:detailed}
\vspace{-.1in}
\end{table}

As verified in Figure~\ref{fig:losses_visualize}, the depth consistency loss of high-quality and low-quality regions differs by nearly an order of magnitude, making it unreasonable to utilize a single loss weight $\lambda_{DC}$. 
In this section, we experimentally verify the effectiveness of the proposed region-aware depth consistency. The pseudo-depth maps are partitioned into high-quality and low-quality regions through online cross-view checking. Then these two regions adopt loss weights $\lambda_{DC}^h$ and $\lambda_{DC}^l$, respectively.
As shown in Table~\ref{tbl:abl_region_aware}, compared with no depth consistency (Overall metric of 0.3420), applying depth consistency independently to high-quality or low-quality regions has a significant positive effect.
For example, when setting low-quality regions' depth consistency loss weight $\lambda_{DC}^l$ to 0.1, the Overall metric improves from 0.3420 to 0.3291.
%
We believe that the key to performance improvement lies in the balance of the accurate and the erroneous pseudo-depth consistencies.
When the loss weight of $\lambda_{DC}^h$ is set to 0.5, and the loss weight of $\lambda_{DC}^l$ is set to 0.1, the model achieves the best performance of 0.3231.
The detailed ablation of loss weights $\lambda_{DC}^h$ and $\lambda_{DC}^l$ is shown in Table~\ref{tbl:detailed}.


\paragraph{Effect of each component.}

\begin{table}[!t]
\centering
\resizebox{0.99\linewidth}{!}{
\begin{tabular}{@{}c|c|c|ccc|c@{}}
\toprule
M. design & A. policy & R.D. & Acc($\downarrow$)    & Comp($\downarrow$)   & Overall($\downarrow$) & GPU [MiB] \\ \midrule
          &     &     &   0.3762 & 0.3203 & 0.3483 & 14100    \\ 
          \midrule
   $\checkmark$  &     &     &   0.3582 & 0.3160 & 0.3371   & 8000 \\ 
     $\checkmark$ & $\checkmark$   &    &  0.3520   &  0.3004   &  0.3262  & 8000 \\
     $\checkmark$  &  $\checkmark$   & $\checkmark$  &  0.3479 & 0.2983 & \textbf{0.3231} &  8000 \\
\bottomrule
\end{tabular}
}
\caption{Ablation study of different components of our proposed self-supervised MVS framework.
``M. design'': memory-efficient design.
``A. policy'': asymmetric view selection policy.
``R.D.'': region-aware depth consistency.
}
\label{tbl:component}
\vspace{-.1in}
\end{table}

To evaluate the performance gain of each component in our approach, we provide factor-by-factor ablation experiments, as shown in Table~\ref{tbl:component}.
We can clearly observe that the three key designs of our approach all yield better reconstruction results, and our framework significantly reduces GPU memory usage.

\subsection{Benchmark Performance}

\paragraph{Evaluation on DTU Dataset.}

We evaluate the depth prediction performance on the DTU test set, and compare with previous state-of-the-art methods in Table~\ref{tbl:sota_dtu}. Our \ours{} architecture achieves the best 
overall score (lower is better) among all end-to-end self-supervised MVS methods. Our model improves the overall score from 0.345 of
RC-MVSNet~\cite{rcmvsnet-2022} to 0.323.
The overall score is also better than the multi-stage self-supervised
approaches, and even most supervised methods. 
Visualizations of our point cloud reconstruction results can be found in Figure~\ref{fig:dtu_visualization}.

\begin{table}[!t]
\centering
\resizebox{0.99\linewidth}{!}{
\begin{tabular}{@{}l|c|ccc@{}}
\toprule
                  & Method & Acc($\downarrow$)    & Comp($\downarrow$)   & Overall($\downarrow$) \\ \midrule
\multirow{8}{*}{Supervised} &    SurfaceNet~\cite{surfacenet_17}    &  0.450   & 1.040    & 0.745     \\ 
                  &  MVSNet~\cite{mvsnet_18}      &  0.396   & 0.527     &  0.462   \\ 
                  &  Cas-MVSNet~\cite{cascademvs_2020}      &  0.325   &    0.385  &  0.355  \\ 
                  &   PatchmatchNet~\cite{patchmatchnet_21}     &  0.427   &   0.277   &    0.352 \\
                  &   CVP-MVSNet~\cite{cvp-mvsnet-20}    & 0.296   & 0.406     &   0.351 \\
                  &  UCSNet~\cite{ucsnet_20}     &   0.338 &  0.349    &  0.344  \\
                  & UniMVSNet~\cite{unimvsnet_2022} & 0.352 & 0.278 & 0.315 \\
                  &   GBi-Net~\cite{gbinet_22}    & 0.315   &  0.262    &  0.289  \\                  
\midrule

\multirow{3}{*}{Multi-Stage Self-Sup.} &   Self\_sup CVP-MVSNet~\cite{YangAL21}     &   0.308  &   0.418   &  0.363 \\ 
                  &    U-MVSNet~\cite{umvsnet_21}    &  0.354    &  \hspace{.03in} 0.3535    &  0.354 \\ 
                  &  KD-MVS~\cite{kd-mvs-22}      &  0.359   &  0.295    &     0.327 \\ 
\midrule

\multirow{7}{*}{E2E Self-Sup.} &   Unsup\_MVSNet~\cite{Khot_2019}     &  0.881   & 1.073     &     0.977 \\
                  &   MVS2~\cite{MVS2_2019}     &  0.760   &  0.515    &  0.637  \\ 
                  &   M3VSNet~\cite{M3VSNET_2021}     &  0.636   &  0.531    &  0.583  \\ 
                  & DS-MVSNet~\cite{DSMVSNet_2022} & 0.374 & 0.347 & 0.361 \\
                  &   JDACS-MS~\cite{jdacs_2021}      &  0.398   &  0.318    &   0.358  \\ 
                  &   RC-MVSNet~\cite{rcmvsnet-2022}     &   0.396  & 0.295    &    0.345 \\ 
                  &   \textbf{\ours{}} (ours) & 0.348  & 0.298 &  \textbf{0.323} \\

\bottomrule
\end{tabular}
}
\caption{Point cloud evaluation results on DTU dataset~\cite{dtu_dataset_2016}. 
The lower is better.
The sections are partitioned into supervised, multi-stage self-supervised, and end-to-end self-supervised, respectively. 
All the results other than ours are from previously published literature.}
\vspace{-.15in}
\label{tbl:sota_dtu}
\end{table}

\paragraph{Evaluation on Tanks\&Temples.}

\begin{table}[!t]
\centering
\resizebox{0.8\linewidth}{!}{
\begin{tabular}{@{}l|c|cc@{}}
\toprule
                  & Method & Mean $\uparrow$ \\ \midrule
\multirow{8}{*}{Supervised} &    MVSNet~\cite{mvsnet_18}    &   43.48  \\
                  &  CIDER~\cite{cider_2020}      &   46.76 \\ 
                  &  PatchmatchNet~\cite{patchmatchnet_21}      &     53.15  \\ 
                  &   CVP-MVSNet~\cite{cvp-mvsnet-20}     &  54.03  \\ 
                  &  UCSNet~\cite{ucsnet_20}      &   54.83  \\ 
                  &  Cas-MVSNet~\cite{cascademvs_2020}      &   56.42   \\ 
                  & AA-RMVSNet~\cite{aa-rmvsnet_2021} & 61.51 \\
                  & UniMVSNet~\cite{unimvsnet_2022} & 64.36  \\
\midrule
\multirow{3}{*}{Multi-Stage Self-Sup.} &   Self\_sup CVP-MVSNet~\cite{YangAL21}     &   46.71  \\
                  &   U-MVSNet~\cite{umvsnet_21}     &   57.15   \\ 
                  &   KD-MVS~\cite{kd-mvs-22}     &  \hspace{-.1in} $^{\ddag}$64.14  \\ 
\midrule
\multirow{6}{*}{E2E Self-Sup.} &   MVS2~\cite{MVS2_2019}     &   37.21   \\
                  &   M3VSNet~\cite{M3VSNET_2021}     &     37.67 \\
                   &   JDACS-MS~\cite{jdacs_2021}     &   45.48  \\
                   & DS-MVSNet~\cite{DSMVSNet_2022} & 54.76 \\
                    &   RC-MVSNet~\cite{rcmvsnet-2022}     &   55.04   \\
                     &  \textbf{\ours{}} (ours)     &  \textbf{58.10}  \\
\bottomrule
\end{tabular}
}
\caption{Point cloud evaluation results on the \textit{intermediate} subsets of Tanks\&Temples dataset~\cite{tanks_2017}. Higher scores are better. 
The Mean is the average score of all scenes. 
$^{\ddag}$ denotes further finetuning on the BlendedMVS~\cite{blendmvs_2020} dataset.
}
\vspace{-.15in}
\label{tbl:tt_intermediate_sota}
\end{table}

\begin{table}[!t]
\centering
\resizebox{0.7\linewidth}{!}{
\begin{tabular}{@{}l|c|c@{}}
\toprule
                  & Method & Mean$\uparrow$  \\ \midrule
\multirow{6}{*}{Supervised} &   CIDER~\cite{cider_2020}     &     23.12  \\ 
                  &    R-MVSNet~\cite{rmvsnet_2019}    &  24.91 \\ 
                  &  CasMVSNet~\cite{cascademvs_2020}      &    31.12  \\ 
                  &   PatchmatchNet~\cite{patchmatchnet_21}     &  32.31 \\ 
                  &    AA-RMVSNet~\cite{aa-rmvsnet_2021}    &   33.53  \\
                  & UniMVSNet~\cite{unimvsnet_2022} & 38.96 \\
\midrule
\multirow{2}{*}{Multi-Stage Self-Sup.} &  U-MVSNet~\cite{umvsnet_21} &  30.97  \\ 
                  &    KD-MVS~\cite{kd-mvs-22}    & \hspace{-.1in} $^\ddag$37.96 \\ \midrule
\multirow{2}{*}{E2E Self-Sup.} &   RC-MVSNet~\cite{rcmvsnet-2022} &  30.82 \\
              & \textbf{\ours{}} (ours)           & \textbf{34.90}  \\ 
\bottomrule
\end{tabular}
}
\caption{Point cloud evaluation results on the \textit{advanced} subset of Tanks\&Temples dataset~\cite{tanks_2017}.
Higher scores are better. The Mean is the average score of all scenes. 
$^{\ddag}$ denotes further finetuning on the BlendedMVS~\cite{blendmvs_2020} dataset.
}
\vspace{-.15in}
\label{tbl:tt_advance_sota}
\end{table}

We train the proposed \ours{} on the DTU training set and test it on the Tanks\&Temples dataset without fine-tuning. We compare our method with state-of-the-art supervised, pseudo-label-based multi-stage self-supervised and end-to-end self-supervised approaches. 
Tables~\ref{tbl:tt_intermediate_sota} and \ref{tbl:tt_advance_sota} show the evaluation results for the intermediate and advanced subsets, respectively.
Our \ours{} achieves the best performance among end-to-end self-supervised methods on both subsets. 
The anonymous evaluation on the leaderboard~\cite{tanks_leaderboard} is named \ours{}. 
The qualitative point cloud reconstruction results are visualized in Figure~\ref{fig:tanks_visualize}.
More detailed comparisons of each scene are reported in our supplementary materials.

\section{Conclusions}

In this work, we propose an efficient E2E self-supervised MVS framework, named \ours{}, which achieves state-of-the-art performance without the need for additional third-party models. The \ours{} model includes three key designs: a memory-efficient structure, an asymmetric view selection policy, and a region-aware depth consistency.

{\small
\bibliographystyle{ieee_fullname}
\bibliography{egbib}
}

\end{document}